\documentclass[10pt,twocolumn,letterpaper]{article}

\usepackage{cvpr} 

\usepackage{adjustbox}

\definecolor{cvprblue}{rgb}{0.21,0.49,0.74}
\usepackage[pagebackref,breaklinks,colorlinks,allcolors=cvprblue]{hyperref}
\usepackage{algorithm}
\usepackage{algorithmic}

\usepackage{tikz}
\usetikzlibrary{decorations.pathreplacing}

\usepackage{xcolor}

\usepackage[table]{xcolor}
\usepackage{booktabs}
\usepackage{multirow}
\usepackage{hhline}
\usepackage{tabularx} 
\usepackage{array}    
\newcommand{\ourmodel}{TF-SSD}
\usepackage{amssymb}

\usepackage{tikz}
\usetikzlibrary{tikzmark}
\definecolor{mygray}{gray}{0.9}

\newcommand{\marked}[1]{#1}  
\newcommand{\markedred}[1]{#1}

\newcommand{\INPUT}{\item[\textbf{Input:}]}
\newcommand{\OUTPUT}{\item[\textbf{Output:}]}
\newcommand{\INITIALIZE}{\item[\textbf{Initialize:}]}
\makeatletter
\renewcommand{\fs@ruled}{%
  \def\@fs@cfont{\bfseries}
  \let\@fs@capt\floatc@ruled
  \def\@fs@pre{\kern2pt\hrule height 1.1pt \kern2pt}
  \def\@fs@post{\kern2pt\hrule height 1.1pt\relax}
  \def\@fs@mid{\kern2pt\hrule height 1.1pt\kern2pt}
  \let\@fs@iftopcapt\iftrue
}
\makeatother

\def\paperID{35959} 
\def\confName{CVPR}
\def\confYear{2026}

\title{TF-SSD: A Strong Pipeline via Synergic Mask Filter for Training-free \\ Co-salient Object Detection}

\author{
Zhijin He\textsuperscript{1}\thanks{Equal contribution.} \quad 
Shuo Jin\textsuperscript{1,2}\footnotemark[1] \quad 
Siyue Yu\textsuperscript{1}\thanks{Corresponding author: siyue.yu02@xjtlu.edu.cn} \quad
Shuwei Wu\textsuperscript{1} \quad  
Bingfeng Zhang\textsuperscript{3} \quad 
Li Yu\textsuperscript{4} \quad 
Jimin Xiao\textsuperscript{1}\\
$^{1}$XJTLU   \quad  
 $^{2}$University of Liverpool  \quad
 $^{3}$China University of Petroleum (East China) \\
 $^{4}$Nanjing University of Information Science and Technology\\
{\tt\small \{Zhijin.He24, Shuwei.Wu24\}@student.xjtlu.edu.cn},
{\tt\small shuo.jin@liverpool.ac.uk},
\\
{\tt\small \{siyue.yu02, jimin.xiao\}@xjtlu.edu.cn},
{\tt\small bingfeng.zhang@upc.edu.cn},
{\tt\small li.yu@nuist.edu.cn}
}

\begin{document}
\maketitle

\begin{abstract}
Co-salient Object Detection (CoSOD) aims to segment salient objects that consistently appear across a group of related images. Despite the notable progress achieved by recent training-based approaches, they still remain constrained by the closed-set datasets and exhibit limited generalization. However, few studies explore the potential of Vision Foundation Models (VFMs) to address CoSOD, which demonstrate a strong generalized ability and robust saliency understanding.
In this paper, we investigate and leverage VFMs for CoSOD, and further propose a novel training-free method, TF-SSD, through the synergy between SAM and DINO. 
Specifically, we first utilize SAM to generate comprehensive raw proposals, which serve as a candidate mask pool. Then, we introduce a quality mask generator to filter out redundant masks, thereby acquiring a refined mask set. Since this generator is built upon SAM, it inherently lacks semantic understanding of saliency. To this end, we adopt an intra-image saliency filter that employs DINO's attention maps to identify visually salient masks within individual images. Moreover, to extend saliency understanding across group images, we propose an inter-image prototype selector, which computes similarity scores among cross-image prototypes to select masks with the highest score. These selected masks serve as final predictions for CoSOD. Extensive experiments show that our TF-SSD outperforms existing methods (\eg, 13.7\% gains over the recent training-free method). Codes are available at \url{https://github.com/hzz-yy/TF-SSD}.
\end{abstract}    
\section{Introduction}
\label{sec:intro}

Co-salient Object Detection (CoSOD) is an emerging computer vision task that extends traditional Salient Object Detection (SOD) to identify common salient objects across multiple related images~\cite{fan2020taking,zhang2020coadnet}. The key challenge is to ensure the models can identify visual consensus across images, instead of relying on class-specific semantic patterns.

Existing state-of-the-art (SOTA) methods follow the training-based paradigm and have achieved impressive performance. 
However, they are mostly trained on closed-set datasets and exhibit limited generalization~\cite{xiao2024zero}.
These problems exhibit a fundamental mismatch: CoSOD requires discovering generalizable visual consensus across objects, while conventional supervised methods often overfit to biased class distributions of the training set.

\begin{figure}[t]
  \centering
  \includegraphics[width=\columnwidth]  {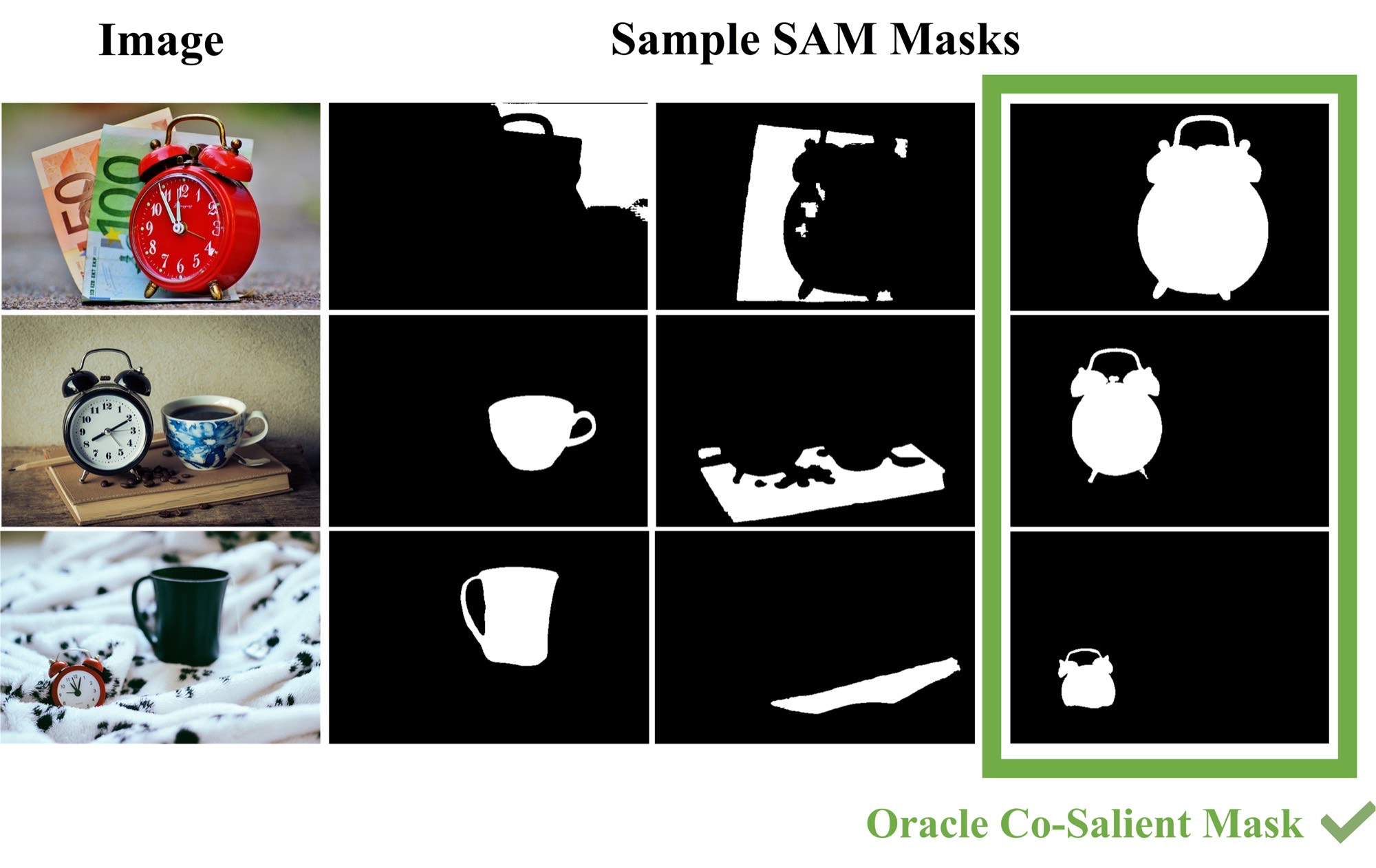}
  \caption{Segmentation results derived from SAM. The oracle masks selected under the ground truth guidance exhibit co-salient representations, which are highlighted in the green rectangle.}
  \label{fig:intro1}
  \vspace{-2mm}
\end{figure}

\begin{figure}[t]
  \centering
  \includegraphics[width=1.0\columnwidth]{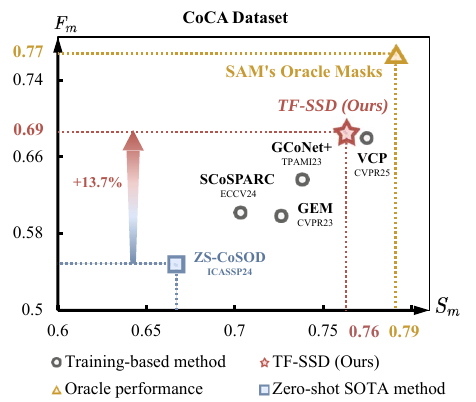}
  \caption{Comparison of our TF-SSD and other CoSOD methods. $F_m$ and $S_m$ denote the F-measure and S-measure, respectively. Our method surpasses training-based methods and achieves 13.7\% improvement over the recent training-free SOTA method.}
  \label{fig:intro2}
  \vspace{-2mm}
\end{figure}

Recent advances in vision foundation models (VFM), such as SAM~\cite{sam}, and DINO~\cite{dino,oquab2023dinov2}, have opened new avenues for various vision tasks. SAM exhibits an impressive generalization ability for image segmentation. As depicted in Fig.~\ref{fig:intro1}, SAM generates numerous masks, which contain co-salient objects with accurate delineation. It motivates us to explore the potential of SAM for CoSOD. Further, we conduct an experiment to quantify SAM's segmentation ability in CoSOD benchmarks. As shown in Fig.~\ref{fig:intro2}, when selecting oracle masks\footnote{We select the best mask result from SAM's top-10 predictions using ground truth (GT) mask guidance, which are shown in Fig.~\ref{fig:intro1}.} from SAM's outputs, the segmentation performance on CoCA benchmark~\cite{zhang2020gradient} outperforms all existing SOTA methods, including fully-supervised approaches~\cite{zhu2023co,li2023discriminative,ge2022tcnet,wu2023co,zhang2022deep}. This upper bound result suggests that SAM’s segmentation capability is sufficient for CoSOD. However, SAM inherently lacks the semantic understanding required to distinguish cross-image relationships and to determine the co-salient predictions. This gap reveals a crucial question: `\textit{How to capture common semantics for identifying accurate co-salient masks from SAM's exhaustive proposals?}' 

To address this, we turn to the self-supervised VFM, DINO~\cite{dino}, which has demonstrated an exceptional ability to extract rich and salient semantic features. Its dense features and attention maps exhibit strong semantic coherence for salient objects~\cite{simeoni2025dinov3,oquab2023dinov2}, which provides accurate semantic understanding that naturally complements SAM's segmentation results. Such a phenomenon enables a mutually complementary paradigm: DINO's semantic guidance can address SAM's stochastic nature in mask generation, while SAM's precise boundary perception enhances DINO's coarse localization.
Building on these insights, we introduce a novel \textbf{T}raining-\textbf{F}ree approach for CoSOD through the \textbf{S}ynergy between \textbf{S}AM and \textbf{D}INO, termed \textbf{TF-SSD}. Our core strategy is to treat SAM's exhaustive mask proposals as a rich candidate mask pool and progressively refine them through complementary guidance of DINO at two levels: intra-image visual saliency and inter-image semantic consistency.

Specifically, our TF-SSD utilizes a progressive pipeline including three components. Since SAM typically generates numerous mask proposals for salient foreground objects, including multiple extremely small and overlapping masks, we first propose a \textbf{Quality Mask Generator (QMG)} to progressively filter out these low-quality and redundant masks based on intrinsic properties, such as area and size, to obtain \marked{the refined mask proposal set}. Nevertheless, QMG can't distinguish the visually salient object within each image. We thus introduce an \textbf{Intra-image Saliency Filter (ISF)}. It leverages DINO's attention maps to identify visually salient masks by evaluating the spatial alignment between attention responses and mask regions. However, these operations are limited to individual images, failing to capture the co-existing relationships inside the group. We further propose an \textbf{Inter-image Prototype Selector (IPS)} for inter-image semantic consistency. In detail, IPS computes similarity scores among inter-image prototypes, which are established by the masks from our ISF and global features from DINO. The masks with the highest score in each image are then selected as the final co-salient predictions for CoSOD.

These components work in concert and effectively boost the performance of CoSOD. As shown in Fig.~\ref{fig:intro2}, our TF-SSD outperforms existing training-based methods and surpasses the recent training-free method up to 13.7\% in F-measure on the CoCA benchmark. In summary, our contributions are concluded as follows:

\begin{itemize}

\item We propose a novel training-free framework, TF-SSD, for CoSOD. It progressively transforms SAM's exhaustive mask proposals into co-salient mask predictions under the guidance of DINO's feature representation.

\item We propose QMG and ISF to filter high-quality mask proposals and detect visually salient masks within each image based on intra-image visual saliency. To facilitate co-salient understanding across images, IPS is introduced to model inter-image prototype relations and select semantically consistent masks as the final CoSOD predictions.


\item Extensive experiments on CoSOD benchmarks show that TF-SSD achieves SOTA performance, which outperforms both training-based and training-free approaches.

\end{itemize}

\section{Related Works}
\label{sec:formatting}

\subsection{Co-salient Object Detection}
CoSOD requires segmenting salient objects that are commonly present across a group of images. Recent advances have achieved significant progress, and existing methods can be categorized into supervised ~\cite{zheng2023memory,zheng2023gconet+,hou2017deeply,li2016deep,vaze2022generalized,xu2023co,le2017co,zhu2023co,li2023discriminative,ge2022tcnet,wu2023co,zhang2022deep}, unsupervised ~\cite{wang2023tokencut,Chakraborty_2024_WACV,amir2021deep,yan2022unsupervised,hsu2018co,liu2021semi,liu2023self}, and zero-shot ~\cite{xiao2024zero} paradigms.

Most supervised methods follow a three-stage paradigm including feature encoding , consensus extraction and dispersion, and final prediction. Some methods actively model both the commonality and the distinction between groups. GCoNet+~\cite{zheng2023gconet+} introduces a group collaborative learning framework, which leverages negative relations between groups to learn more discriminative features. VCP~\cite{wang2025visual} proposes embedding the extracted visual consensus into prompts, which forms a prompt tuning approach with minimized tunable parameters.

To address annotation dependency, unsupervised solutions~\cite{wang2023tokencut,Chakraborty_2024_WACV,amir2021deep} are explored to learn co-salient objects solely from unlabeled data. SCoSPARC~\cite{chakraborty2024self} introduces a two-stage self-supervised model involving a novel confidence-based adaptive threshold, which enhances cross-image feature correspondence. Moreover, the zero-shot paradigm has advanced with the emergence of vision foundation models. ZS-CoSOD~\cite{xiao2024zero} pioneers this research direction by integrating powerful foundation models to generate custom group prompts that represent the co-salient objects to guide SAM in predicting results for CoSOD in a training-free manner.
PAP-SAM~\cite{yu2025pap} further introduces a global-local prior adaptive perception strategy to enhance SAM for CoSOD. However, few studies have explored how to endow the coarse segmentation results derived from SAM with co-salient semantic understanding.

\subsection{Vision Foundation Model}
Building upon large-scale high-quality datasets~\cite{imagenet,oquab2023dinov2,shao2019objects365,sam,xue2025mmrc}, learning generalized visual representations has become a popular paradigm in computer vision. Vision Foundation Models (VFMs) aim to undergo pre-training on large-scale datasets to learn general visual representations, which can generalize to diverse downstream tasks, including image-level recognition~\cite{clip,li2022blip}, pixel-level perception~\cite{maskclip,vader,jin2025feature}, and video restoration~\cite{liu2022temporal,jin2023kernel}. 

One research branch of VFM is self-supervised models~\cite{mae,zhou2021ibot}, which aim to learn general visual features solely from images. Among these, the DINO series~\cite{dino,oquab2023dinov2,simeoni2025dinov3} have shown impressive capabilities to capture explicit details in their dense features, while its attention map can localize the precise foreground object. Another research branch of VFM is the SAM series~\cite{sam,ravi2024sam2}, which are generally pre-trained on a large-scale annotated training set in a fully-supervised paradigm. SAM~\cite{sam} has demonstrated impressive zero-shot, class-agnostic segmentation ability for general image segmentation tasks. To address CoSOD in a training-free manner, we turn to VFMs that generate comprehensive mask proposals and filter the co-salient ones through the synergy between SAM and DINO.

\section{Methodology}
The overview pipeline of our TF-SSD is presented in Sec.~\ref{sec:ov}, which comprises three key components: 1)  Quality Mask Generator (QMG) in Sec.~\ref{sec:qmg}; 2) Intra-image Saliency Filter (ISF) in Sec.~\ref{sec:isf}; and 3) Inter-image Prototype Selector (IPS) in Sec.~\ref{sec:IPS}.

\begin{figure*}[!t]
  \centering
  \includegraphics[width=0.95\textwidth]{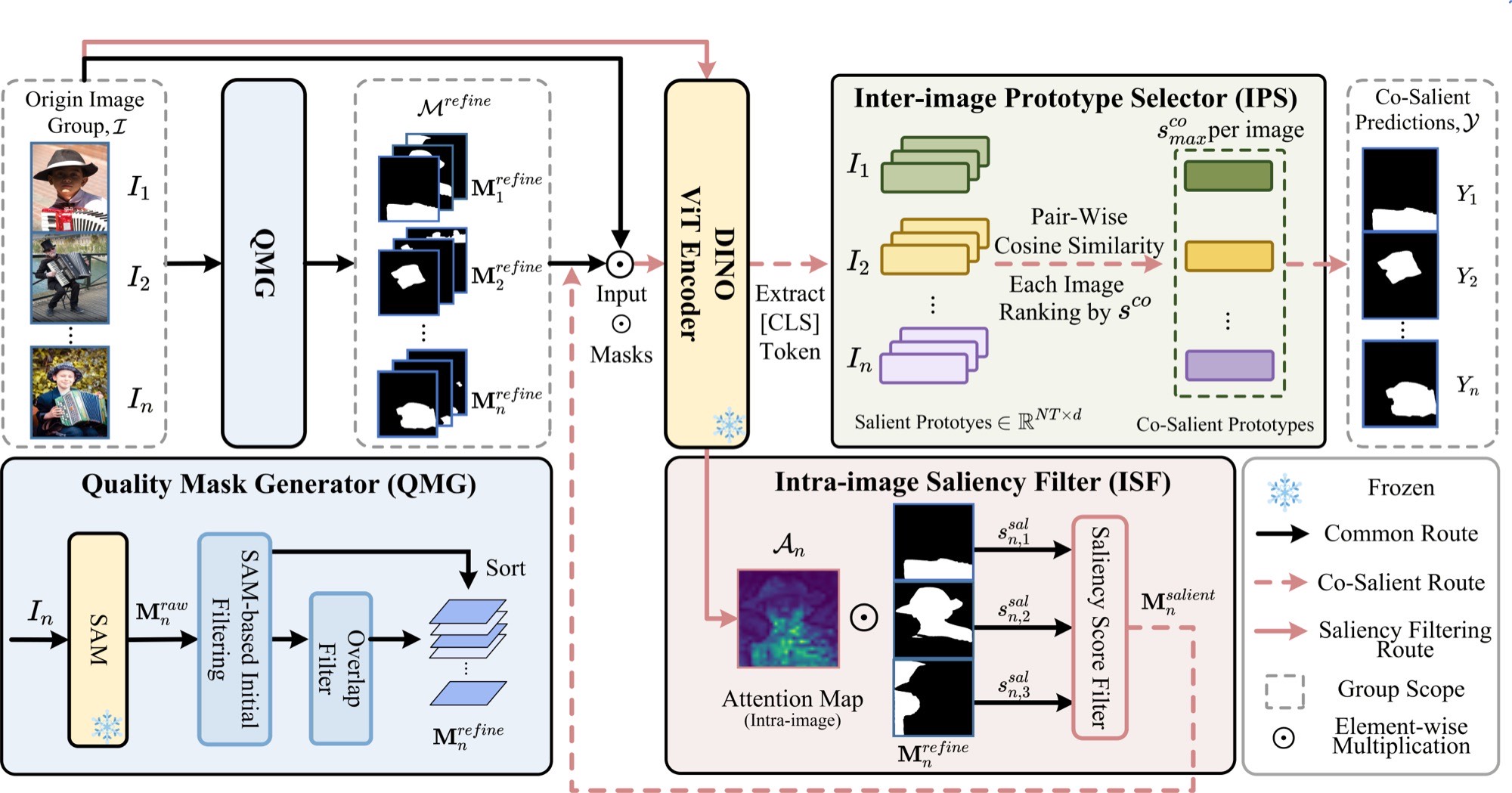}  
  \caption{Overview pipeline of our TF-SSD. It contains four components: a Quality Mask Generator (QMG) that filters out exhaustive candidate masks from SAM, a DINO encoder for salient attention and semantic prototype extraction, an Intra-image Saliency Filter (ISF) for visually salient purification within a single image, and an Inter-image Prototype Selector (IPS) that builds the saliency relationship across group images to discover the co-salient masks as the final predictions for CoSOD. For clarity, our notations are defined as: $m_{n,t}$ denotes an individual mask, $\textbf{M}_n$ denotes the mask set of image $n$ (e.g., $\textbf{M}_n^{raw}$, $\textbf{M}_n^{refine}$).}
  \label{fig:pipeline}
\end{figure*}

\subsection{Overview}
\label{sec:ov}
Given a group of $N$ images $\mathcal{I} = \{I_1, I_2, \ldots, I_n\}$ that contain salient objects of a common category across the group. CoSOD seeks to identify co-salient objects and generate corresponding segmentation masks $\mathcal{Y} = \{Y_n\}_{n=1}^{N}$.
Our framework operates through a progressive pipeline that narrows SAM's masks to co-salient predictions for CoSOD, as shown in Fig.~\ref{fig:pipeline}. The overall pipeline is as follows: 

\begin{enumerate}[1)]
\item For each image $I_n \in \mathcal{I}$, SAM is first used to generate comprehensive segmentation proposals ${\textbf{M}}_n^{raw}$. Then, our QMG employs a multi-stage strategy to filter out redundant masks, obtaining refined masks ${\textbf{M}}_n^{refine}$.

\item Next, owing to the absence of visual saliency perception in ${\textbf{M}}_n^{refine}$ within a single image, our ISF employs attention maps from DINO to highlight visually salient regions. It filters masks with low saliency scores using an adaptive threshold, yielding the filtered masks ${\textbf{M}}_n^{salient}$.

\item Subsequently, since most images contain multiple salient objects, our IPS extracts prototypes from each mask $m_{n,t} \in {\textbf{M}}_n^{salient}$ to compute the co-salient scores. Finally, masks with the highest co-salient score are selected as the final prediction $\mathcal{Y}$ for CoSOD.

\end{enumerate}

\subsection{Quality Mask Generator}
\label{sec:qmg}

Quality Mask Generator (QMG) serves as the foundation of our framework, which transforms SAM's exhaustive mask proposals into a refined mask proposal set, prepared for co-salient detection. Given an input image $I_n$, SAM initially generates a raw mask set ${\textbf{M}}_{n}^{raw} = \{m^{raw}_{n,1}, m^{raw}_{n,2}, \ldots, m^{raw}_{n,T_o}\}$ where $T_o$ denotes the mask number of each image. However, the nature of the SAM generation process introduces significant limitations: most masks correspond to unrelated objects and redundant segments of the same object. 

To address this, QMG progressively eliminates trivial, overlapping, and excessively sized masks. This process operates through: SAM-based initial filtering, overlap filtering, and  quality assessment, summarized in Algorithm~\ref{alg:qmg} for a clear view. 

\noindent\textbf{SAM-based Initial Filtering.}
First, ${\textbf{M}}_{n}^{raw}$ contains many masks with excessively small area of the segment, termed trivial masks. We posit that these masks are not associated with salient objects. To discard these trivial masks, we compute an area ratio $r_{n,t}^{area}$, which denotes the proportion of the image covered by the mask, formulated as: 
\begin{equation}
r_{n,t}^{area} = \frac{|m_{n,t}^{raw}|}{H \times W},
\label{eq:area_ratio}
\end{equation}
where $|m_{n,t}|$ denotes the foreground spatial area of the mask $m_{n,t}^{raw}$, $H$ and $W$ denotes the image height and width. $r_{n,t}^{area}$ is used to filter out trivial masks, establishing the coarse mask set ${\textbf{M}}_n^{coarse}$, as formulated:
\begin{equation}
\textbf{M}_n^{coarse} = \{\, m_{n,t}^{raw}\mid r_{n,t}^{area} \ge \tau_{area} \,\},
\label{eq:initial}
\end{equation}
where $\tau_{area}$ is the area threshold.

\noindent\textbf{Overlap Filtering.} Besides the above trivial masks, there also exist multiple overlapping segments. Therefore, an overlap filtering is introduced to further purify ${\textbf{M}}_n^{coarse}$. 

We design the overlap ratio $\rho_{i \rightarrow j}$ for each mask pair:
\begin{equation}
\label{eq:re}
\rho_{i \rightarrow j} = \frac{|m_{n,i} \cap m_{n,j}|}{|m_{n,j}|},
\end{equation}
where $m_{n,i}$ and $m_{n,j}$ denote paired masks. To preserve larger masks, we calculate the overlap ratio of the candidate mask from  ${\textbf{M}}_n^{coarse}$ in ascending order of mask area. If the current mask has a high overlap ratio ($\rho_{i \rightarrow j} \geq \tau_{con}$, $\tau_{con}$ is the overlap threshold) with any larger mask, it is removed. The rest masks are formed as the purified set $\textbf{M}_n^{purified}$.


\noindent\textbf{Quality Assessment.}
{While the preceding area-based filtering is can filter most extreme small and overlapping masks, there are still plenty of noisy masks in ${\textbf{M}}_n^{purified}$.
Thus, we further leverage the IoU score $\text{IoU}_{n,t}^{pred}$ to filter the trivial masks as an auxiliary. Note that SAM predicts the IoU score to measure the prediction quality. We compute a quality-based metric by combining the IoU score and our area ratio.

First, given that salient objects typically have moderate sizes rather than extreme ones, we define an area score $S_{n,t}^{area}$ to assign lower scores to masks with excessive sizes:

\begin{equation}
S_{n,t}^{area} = \begin{cases} 
1.0 & \text{if } r_{min} \leq r_{n,t}^{area} \leq r_{max}, \\
r_{n,t}^{area} / r_{min} & \text{if } r_{n,t}^{area} < r_{min}, \\
\mathcal{P}(r_{n,t}^{area}) & \text{if } r_{n,t}^{area} > r_{max},
\end{cases}
\label{eq:area_score}
\end{equation}
where $\mathcal{P}(r_{n,t}^{area}) = \max(\sigma, 1.0 - (r_{n,t}^{area} - r_{max}) \times \gamma)$ is a penalty function with hyper-parameters $\sigma$ and $\gamma$. It progressively assigns lower scores as the area increases to suppress masks that exceed the ideal size. Eq.~\ref{eq:area_score} assigns the highest score to masks belonging to the ideal size $[r_{min}, r_{max}]$.

Then, we integrate $S_{n,t}^{area}$ with $\text{IoU}_{n,t}^{pred}$, to compute a balanced quality score $S_{n,t}^{ba}$ , as
\begin{equation}
S_{n,t}^{ba} = \alpha \cdot \text{IoU}_{n,t}^{pred} + \beta \cdot S_{n,t}^{area},
\label{eq:balanced_score}
\end{equation}
where $\alpha$ and $\beta$ are weighted factors to balance prediction confidence and size preferences. A higher $S_{n,t}^{ba}$ means the quality of the mask is better.

Finally, according to the quality score, the top $T_r$ masks from ${\textbf{M}}_n^{purified}$ are selected to constitute the QMG's final output  ${\textbf{M}}_n^{refined}=\{m_{n,t}^{refined}\}_{t=1}^{T_r}$.

\begin{algorithm}[t]
\caption{Quality Mask Generator (QMG)}
\label{alg:qmg}
\begin{algorithmic}
\INPUT A group of images $\mathcal{I} = \{I_n\}_{n=1}^N$
\OUTPUT Refined mask sets $\mathcal{M}^{refine}=\{{\textbf{M}}_n^{refine}\}_{n=1}^{N}$
\INITIALIZE ${\textbf{M}}_n^{coarse}$, ${\textbf{M}}_n^{purified}$, ${\textbf{M}}_n^{scored}$, ${\textbf{M}}_n^{refine}$, ${\textbf{M}}_{n}^{raw}$
\STATE\textbf{for } $I_n \in \mathcal{I}$:
\STATE\tikzmark{start}\quad\quad ${\textbf{M}}_n^{raw}\leftarrow \text{SAM}(I_n)$ 
\STATE \quad\quad\textbf{Stage.1}
\STATE\tikzmark{start1}\quad\quad\quad\quad \textbf{for} $m_{n,t}$ \textbf{in} ${\textbf{M}}_n^{raw}$:
\STATE \quad\quad\quad\quad\quad Compute area ratio $r_{n,t}^{area}$ (Eq.~\ref{eq:area_ratio})
\STATE \quad\quad\quad\quad\quad \textbf{if} $r_{n,t}^{area} \ge \tau_{area}$:
\STATE\tikzmark{end1}\quad\quad\quad\quad\quad\quad Add $m_{n,t}$ to ${\textbf{M}}_i^{coarse}$ 
\STATE \quad\quad\textbf{end}
\STATE \quad \quad \textbf{Stage.2}
\STATE\tikzmark{start2}\quad\quad\quad\quad Sort ${\textbf{M}}_n^{coarse}$ by mask area (descending)
\STATE \quad\quad\quad\quad\quad \textbf{for} $m_{n,i}$, $m_{n,j}$ \textbf{in} ${\textbf{M}}_n^{coarse}$:
\STATE \quad\quad\quad\quad\quad\quad Calculate $\rho_{i \rightarrow j}$ (Eq.~\ref{eq:re})
\STATE \quad\quad\quad\quad\quad\quad \textbf{if} $\rho_{i \rightarrow j} <\tau_{con}$ :
\STATE\tikzmark{end2}\quad\quad\quad\quad\quad\quad\quad Add $m_{n,i}$ to ${\textbf{M}}_n^{purified}$ 
\STATE \quad\quad\textbf{end}
\STATE \quad\quad\textbf{Stage.3}
\STATE\tikzmark{start3}\quad\quad\quad\quad \textbf{for} $m_{n,t}$ \textbf{in} ${\textbf{M}}_n^{purified}$:
\STATE \quad\quad\quad\quad\quad Compute area score $S_{n,t}^{area}$ (Eq.~\ref{eq:area_score})
\STATE \quad\quad\quad\quad\quad Compute balanced score $S_{n,t}^{ba}$~(Eq.~\ref{eq:balanced_score})
\STATE \quad\quad\quad\quad Sort ${\textbf{M}}_n^{purified}$ by $S_{n,t}^{ba}$
\STATE\tikzmark{end3}\quad\quad\quad\quad Select top $T_r$ masks to form ${\textbf{M}}_n^{refined}$
\STATE \quad\quad\textbf{end}
\STATE\tikzmark{end}\quad\quad Add $\mathcal{\textbf{M}}_n^{refined}$ to $\mathcal{\textbf{M}}^{refined}$
\STATE \textbf{return} $\mathcal{\mathcal{M}}^{refined}$
\end{algorithmic}
\nointerlineskip
\begin{tikzpicture}[remember picture, overlay]
\draw[line width=0.6pt] 
    ([xshift=2mm, yshift=2mm]pic cs:start) -- ([xshift=2mm, yshift=-2mm]pic cs:end);
\draw[line width=0.6pt] 
    ([xshift=10mm, yshift=2mm]pic cs:start1) -- ([xshift=10mm, yshift=-2mm]pic cs:end1);
\draw[line width=0.6pt] 
    ([xshift=10mm, yshift=2mm]pic cs:start2) -- ([xshift=10mm, yshift=-2mm]pic cs:end2);
\draw[line width=0.6pt] 
    ([xshift=10mm, yshift=2mm]pic cs:start3) -- ([xshift=10mm, yshift=-2mm]pic cs:end3);
\end{tikzpicture}
\end{algorithm}

\subsection{Intra-image Saliency Filter}
\label{sec:isf}
While QMG produces refined segmentations ${\textbf{M}}_n^{refine}$, it lacks salient semantic awareness within images. In contrast, recent works have demonstrated that attention maps from DINO naturally highlight salient objects without explicit supervision. Hence, we propose an Intra-image Saliency Filter (ISF), which leverages DINO's attention representation to select masks corresponding to salient regions.

Specifically, given an image $I_n$, we can extract the attention map $\mathcal{A}_n \in [0,1]$ of the [CLS] token from DINO’s ViT encoder, which is reshaped to $\mathcal{A}_n \in \mathbb{R}^{h \times w}$, where $h$ and $w$ are the height and width. $\mathcal{A}_n$ can highlight visually salient regions with higher response intensity, as shown in the second diagram of the first row of Fig.~\ref{fig:intra_saliency}.

As a result, for each mask proposal $m^{refine}_{n,t} \in {\textbf{M}}_n^{refine}$, we compute its saliency score $s_{n,t}^{sal}$ based on the attention response within the masked region, formulated as:
\begin{equation}
s_{n,t}^{sal} = \frac{1}{|m_{n,t}^{refine}|} \sum \mathcal{A}_n(x,y) \odot m_{n,t}^{refine}(x,y),
\label{eq:saliency_score}
\end{equation}
where $|m_{n,t}^{refine}|$ denotes the spatial area of $m_{n,t}^{refine}$, $(x,y)$ represents spatial location, and $\odot$ denotes the element-wise multiplication. $s_{n,t}^{sal}$ reflects how well the mask region aligns with the salient representation, which is shown in the third row of Fig.~\ref{fig:intra_saliency}. A higher $s_{n,t}^{sal}$ means that the alignment is better and thus the corresponding mask tends to be a salient mask. We select the Top $T$ masks ${\textbf{M}}_n^{refine}$ based on the saliency score to form the salient mask set $\textbf{M}_n^{salient}=\{m_{n,t}^{salient}\}_{t=1}^T$.

\begin{figure}[t]
  \centering
  \includegraphics[width=1.0\columnwidth]{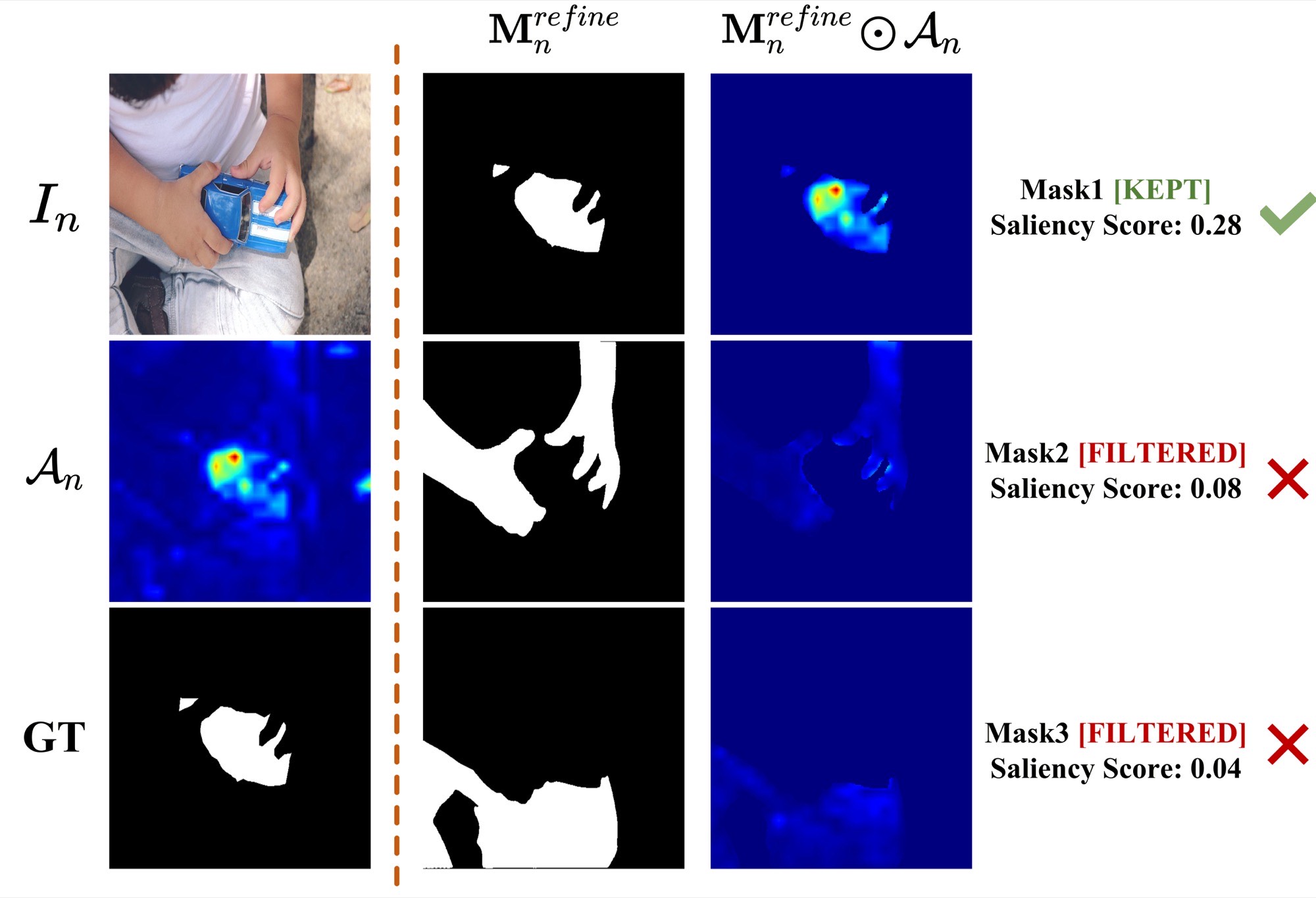}
  \caption{Illustration of Intra-image Saliency Filter. \textbf{Row 1}: Original image, self-attention map of DINO, and ground truth. \textbf{Row 2}: Three mask proposals generated by QMG. \textbf{Row 3}: Attention response obtained by element-wise multiplication of each mask with the attention map. Mask 1 achieves a high saliency score due to strong overlap between salient regions and the mask proposal, while Masks 2 and 3 receive low scores due to non-overlap.}
  \label{fig:intra_saliency}
\end{figure}

\subsection{Inter-image Prototype Selector}
\label{sec:IPS}
While ISF successfully identifies salient regions within individual images, it cannot determine which of these salient objects are co-salient across the entire image group. Therefore, we propose an Inter-image Prototype Selector (IPS) that establishes semantic prototypes for each mask in $\textbf{M}_n^{salient}$ using DINO and matches their consistency across the image group to identify the co-salient masks.

\begin{figure*}[!t]
  \centering
  \includegraphics[width=0.95\textwidth]{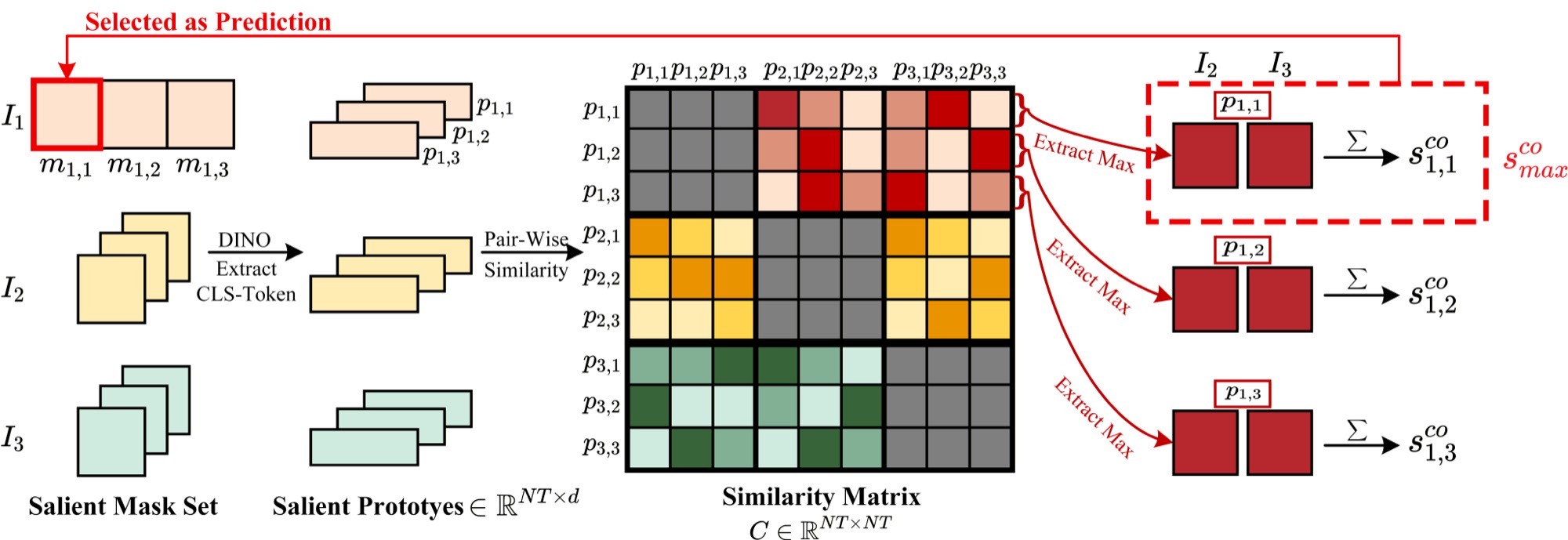}  
  \caption{\markedred{Illustration of Inter-image Prototype Selector (IPS) with an example of N=3 images and T=3 masks per image. A pairwise similarity matrix is constructed using mask prototypes, where different colors denote prototypes from different images. For each candidate (e.g., $p_{1,1}$ of image $I_1$), we select its maximum score (deep red cells) against the candidates from each of the other images ($I_2$ and $I_3$). These maximum scores are summed as the total co-saliency score $s^{co}_{max}$. The mask with the highest $s^{co}_{max}$ is selected as the final prediction.}}
  \label{fig:ips}
\end{figure*}

\marked{The total process is demonstrated in Fig.~\ref{fig:ips}}, for each mask $m_{n,t} \in M_n^{salient}$, we first extract its prototype $p_{n,t}$ using a feature extraction function $\mathcal{F}$:
\begin{equation}
p_{n,t} = \mathcal{F}(I_n \odot m_{n,t}),
\label{eq:prototype}
\end{equation}
where $\mathcal{F}(\cdot)$ means extracting the [CLS] token from DINO's ViT encoder, and $p_{n,t} \in \mathbb{R}^{1 \times d}$ ($d$ is the channel size). This process generates a set of prototypes $P \in \mathbb{R}^{NT \times d}$, where $NT$ means $N$ image in the given group and each has $T$ salient masks derived from our ISF.

Then, the pair-wise cosine similarity matrix $C\in \mathbb{R}^{NT \times NT}$is computed by:
\begin{equation}
    C = \frac{PP^\top}{\left \|P  \right \| ^2}.
\end{equation}
Next, we  reshape the similarity matrix into $C \in \mathbb{R}^{NT\times N \times T}$. However, the similarity score within the same image can be ignored because it's not related to cross-image consistency. Thus, the size of the similarity matrix will become $C \in \mathbb{R}^{NT\times( N-1) \times T}$.

Then, we select the highest similarity score in each image except the target image to get the $N-1$ highest similarity scores for each prototype by:
\begin{equation}
    C^{N-1_{max}} = \max_{i=\{1\cdots T\}}  C[:,:,i],
    \label{eq:maxsimilarityscore}
\end{equation}
where $C^{N-1_{max}} \in \mathbb{R}^{NT \times (N-1)}$.  Afterwards, we derive the co-salient score $s^{co} \in \mathbb{R}^{NT}$ of each prototype by taking the sum of the $N-1$ highest similarity scores, which can be formulated as:
\begin{equation}
    s^{co} = \sum_{n=1}^{N-1} C^{N-1_{max}}[:,n].
\end{equation}
Then, we reshape the co-salient score into $s^{co} \in \mathbb{R}^{N \times T}$, and choose the index of the  prototype with the maximum co-salient score for each image:
\begin{equation}
    s^{co}_{max} = \max_{t=\{1\cdots T\} }s^{co}[:, t],
\end{equation}
\begin{equation}
    index = ind(s^{co}_{max}),
    \label{eq:index}
\end{equation}
where $ind(\cdot)$ means taking the index of the prototype with the maximum co-salient score.

Finally, the mask with the $index$ derived from Eq.~\ref{eq:index} is selected as the final CoSOD prediction for each image. Through this progressive pipeline, our TF-SSD effectively transforms exhaustive mask proposals from SAM into precise co-salient masks across group images.

\section{Experiments}
\begin{table*}[t!]
\footnotesize
\renewcommand{\arraystretch}{1.0}
\setlength\tabcolsep{2.0pt}
\caption{\textbf{Quantitative comparisons between our \ourmodel{}~and other methods.} ``$\uparrow$'' (``$\downarrow$'') means that the higher (lower) is better. `Type' denotes the various supervision settings, where `S' denotes the supervised setting, `U' denotes the unsupervised setting, and `TF' denotes the training-free setting. Best results among training-free (TF) methods are marked \textbf{bold}.} 
\begin{adjustbox}{width=1.0\linewidth,center,valign=t}
\begin{tabular}{l||l|c|cccc|cccc|cccc}
\hline
\rule{0pt}{3mm} &  &  & \multicolumn{4}{c|}{CoCA~\cite{coca}} & \multicolumn{4}{c|}{CoSal2015~\cite{zhang2015co}} & \multicolumn{4}{c}{CoSOD3k~\cite{fan2021re}} \\
Method & Pub. \& Year & Type & MAE $\downarrow$ & $F_\beta^\text{ max} \uparrow$ & $E_{\xi}^\text{max} \uparrow$ & $S_\alpha \uparrow$ & MAE $\downarrow$ & $F_\beta^\text{ max} \uparrow$ & $E_{\xi}^\text{max} \uparrow$ & $S_\alpha \uparrow$ & MAE $\downarrow$ & $F_\beta^\text{ max} \uparrow$ & $E_{\xi}^\text{max} \uparrow$ & $S_\alpha \uparrow$ \\
\hline
\rule{0pt}{3mm}{DCFM}~\cite{yu2022democracy} & CVPR 2022 & S & 0.085 & 0.598 & 0.783 & 0.710 & 0.067 & 0.856 & 0.892 & 0.838 & 0.067 & 0.805 & 0.874 & 0.810 \\
{CoRP}~\cite{zhu2023co} & TPAMI 2023 & S & 0.103 & 0.597 & 0.769 & 0.715 & 0.055 & 0.882 & 0.912 & 0.867 & 0.060 & 0.827 & 0.890 & 0.838 \\
{GCoNet+}~\cite{zheng2023gconet+} & TPAMI 2023 & S & 0.081 & 0.637 & 0.814 & 0.738 & 0.056 & 0.891 & 0.924 & 0.881 & 0.062 & 0.834 & 0.901 & 0.843 \\
{CONDA}~\cite{li2024conda} & ECCV 2024 & S & 0.089 & 0.685 & 0.839 & 0.763 & 0.045 & 0.908 & 0.944 & 0.900 & 0.056 & 0.853 & 0.911 & 0.862 \\
{VCP}~\cite{wang2025visual} & CVPR 2025 & S & 0.069 & 0.680 & 0.829 & 0.774 & 0.037 & 0.920 & 0.944 & 0.911 & 0.049 & 0.868 & 0.868 & 0.874 \\
\hline
\rule{0pt}{3mm}{TokenCut}~\cite{wang2023tokencut} & CVPR 2022 & U & 0.167 & 0.467 & 0.704 & 0.627 & 0.139 & 0.805 & 0.857 & 0.793 & 0.151 & 0.720 & 0.811 & 0.744 \\
{DVFDVD}~\cite{amir2021deep} & ECCVW 2022 & U & 0.223 & 0.422 & 0.592 & 0.592 & 0.092 & 0.777 & 0.842 & 0.809 & 0.104 & 0.722 & 0.819 & 0.773 \\
{US-CoSOD}~\cite{Chakraborty_2024_WACV} & WACV 2024 & U & 0.116 & 0.546 & 0.743 & 0.672 & 0.070 & 0.845 & 0.886 & 0.840 & 0.076 & 0.779 & 0.861 & 0.801 \\
{SCoSPARC}~\cite{chakraborty2024self} & ECCV 2024 & U & 0.092 & 0.614 & 0.782 & 0.711 & 0.062 & 0.869 & 0.905 & 0.851 & 0.064 & 0.827 & 0.889 & 0.823 \\
\hline

\rule{0pt}{3mm}{ZS-CoSOD}~\cite{xiao2024zero} & ICASSP 2024 & TF & 0.115 & 0.549 & - & 0.667 & 0.101 & 0.799 & - & 0.785 & 0.117 & 0.691 & - & 0.723 \\
\rowcolor{mygray}
\textbf{{\ourmodel{}}~(Ours)}	& Submission & TF & \textbf{0.077} & \textbf{0.686} & \textbf{0.815} & \textbf{0.763} & \textbf{0.089} & \textbf{0.899} & \textbf{0.926} & \textbf{0.890} & \textbf{0.090} & \textbf{0.860} & \textbf{0.908} & \textbf{0.861} \\
\hline
\end{tabular}
\end{adjustbox}
\label{table:main}
\end{table*}

\begin{table*}
\begin{center}
\footnotesize
\renewcommand{\arraystretch}{1.0}
\setlength\tabcolsep{1.5pt}
\caption{\textbf{Ablation study of the proposed components in TF-SSD on three benchmark datasets.} QMG-1, QMG-2, and QMG-3 denote the three stages of our QMG, respectively. The final row represents our full framework. Best results are marked \textbf{bold}.}
\label{tab:ablation_modules}
\resizebox{\textwidth}{!}{%
\begin{tabular}{c|ccccc||cccc|cccc|cccc}
\hline
& \multicolumn{5}{c||}{Component}  & \multicolumn{4}{c|}{CoCA~\cite{zhang2020gradient}} & \multicolumn{4}{c|}{CoSal2015~\cite{zhang2015co}} & \multicolumn{4}{c}{CoSOD3k~\cite{fan2021re}} \\
ID &  \hspace{1.25mm} QMG-1 & QMG-2 & QMG-3 &  ISF  &  IPS  & $MAE \downarrow$  & $F_{\beta}^{max}\uparrow$ & $E_{\xi}^{max}\uparrow$ & $S_{\alpha}\uparrow$ & $MAE \downarrow$  & $F_{\beta}^{max}\uparrow$ & $E_{\xi}^{max}\uparrow$ & $S_{\alpha}\uparrow$ & $MAE \downarrow$  & $F_{\beta}^{max}\uparrow$ & $E_{\xi}^{max}\uparrow$ & $S_{\alpha}\uparrow$ \\
\hline
1 & \checkmark &  &  & & \checkmark & 0.171 &0.447 &0.662 &0.578 &0.133 &0.680 &0.779 &0.704 &0.139 &0.610 &0.752 &0.667  \\
2 & \checkmark & \checkmark &  & & \checkmark &0.121&0.497&0.747&0.650&0.095&0.815&0.875&0.799&0.099&0.744&0.839 & 0.754 \\
3 & \checkmark & \checkmark & \checkmark &  & \checkmark &0.119&0.522&0.765&0.634&0.092&0.850&0.900&0.818&0.099&0.785&0.862&0.786\\
\hline
5 & \checkmark & \checkmark & \checkmark & \checkmark & \checkmark &  \textbf{0.077} & \textbf{0.686} & \textbf{0.815} & \textbf{0.763} &
\textbf{0.089} & \textbf{0.899} & \textbf{0.926} & \textbf{0.890} &
\textbf{0.090} & \textbf{0.860} & \textbf{0.908} & \textbf{0.861} \\
\hline
\end{tabular}
}
\end{center}
\end{table*}

\subsection{Datasets and Evaluation Metrics}

We conduct comprehensive experiments to verify the effectiveness on three widely-used benchmarks: CoCA~\cite{zhang2020gradient}, CoSal2015~\cite{zhang2015co}, and CoSOD3k~\cite{fan2021re}. For evaluation, we employ several widely-used metrics:
1) F-measure ($F_\beta^\text{ max}$), which represents the harmonic mean of precision and recall values, calculated using a self-adaptive threshold.
2) S-measure ($S_\alpha$), which is utilized to assess the spatial structural similarities of saliency maps.
3) Mean Absolute Error (MAE) that quantifies the average L1 distance between GT and predictions.
4) $E_{\xi}^\text{max}$, a cognitive-based metric to capture both global statistics and local pixel-level similarity.

\subsection{Implementation Details}
\noindent\textbf{Models and Environment.}
Our training-free framework is built upon two foundational models: SAM and DINO. Specifically, we utilize SAM~\cite{sam} with a ViT-H backbone for mask generation and DINO~\cite{dino,oquab2023dinov2,simeoni2025dinov3} with a ViT-B/8 architecture for feature and attention map extraction, respectively. All experiments are conducted on a single NVIDIA 4090 GPU. For prototype extraction with DINO, the masked image regions are resized to 224x224 pixels.

\noindent\textbf{Parameters.}
Our progressive pipeline involves several key hyperparameters that are discussed as follows:
\begin{itemize}
\item \textbf{Quality Mask Generator (QMG):}
The initial area filtering threshold $\tau_{area}$ is set to 0.01. For overlap filtering, the threshold $\tau_{con}$ is set to 0.85. In the quality assessment stage, the ideal area range $[r_{min}, r_{max}]$ is set to [0.15, 0.7], with the parameters $\sigma=0.7$ and $\gamma=1.5$. The weights of balanced quality score are $\alpha=0.7$ and $\beta=0.3$. We select the top $T_r=10$ masks for next stage.
\item \textbf{Intra-image Saliency Filter (ISF):}
Based on the saliency score computed from DINO's attention map, we select the top $T=6$ masks as salient ones for each image.
\end{itemize}
 More details are provided in the supplementary materials.

\subsection{Comparisons with State-of-the-art Methods}
We compare our TF-SSD with existing SOTA methods under different settings, including fully-supervised methods: DCFM~\cite{yu2022democracy}, CoRP~\cite{zhu2023co}, GCoNet+~\cite{zheng2023gconet+}, CONDA~\cite{li2024conda}, VCP~\cite{wang2025visual}, unsupervised methods: TokenCut~\cite{wang2023tokencut}, DVFDVD~\cite{amir2021deep}, US-CoSOD~\cite{Chakraborty_2024_WACV}, SCoSPARC~\cite{chakraborty2024self}, and a training-free approach, ZS-CoSOD~\cite{xiao2024zero}.

\noindent
\textbf{Quantitive comparison}.
Tab.~\ref{table:main} illustrates comparison results. It's observed that our TF-SSD outperforms all existing SOTA methods. Compared with the recent training-free method, ZS-CoSOD~\cite{xiao2024zero}, our TF-SSD achieves gains of 3.8\% for MAE, 13.7\% for $F_\beta^\text{max}$, and 9.6\% for $S_\alpha$ on the CoCA benchmark. For other challenging real-world benchmarks, \ie, CoSal2015 and CoSOD3k, our TF-SSD also gets promising performance. For example, TF-SSD achieves a gain of 1.2\% for MAE, 10.0\% for $F_\beta^\text{max}$, and 10.5\% for $S_\alpha$ on the CoSal2015 benchmark. In addition, our method achieves comparable performance to supervised learning methods and outperforms unsupervised learning methods. For instance, TF-SSD achieves a gain of 3.0\% for $F_\beta^\text{max}$ and 3.9\% for $S_\alpha$ on the CoSal2015 benchmark.

\begin{figure*}[!t]
  \centering
  \includegraphics[width=0.95\textwidth]{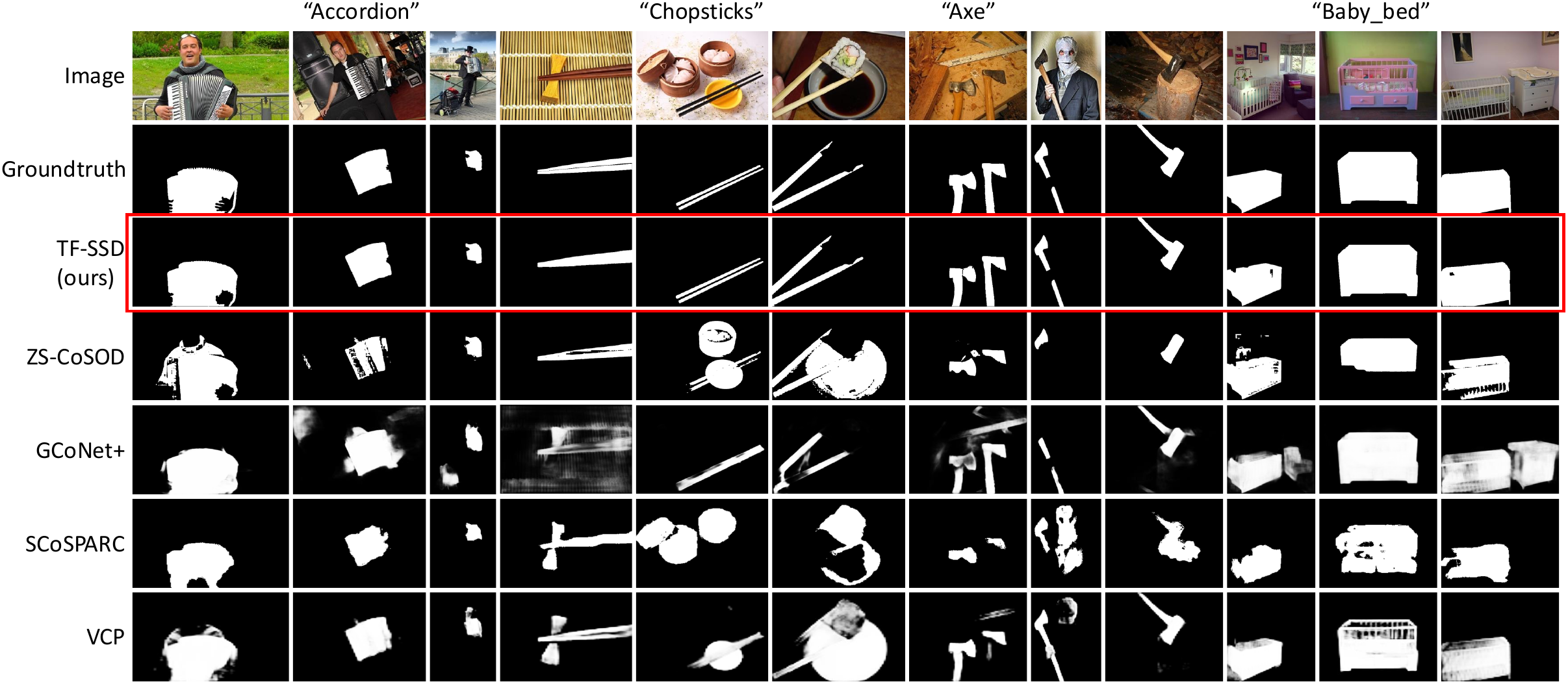}  
  \caption{Qualitative comparison. We compare the result visualization between our method and previous SOTA methods. Among them, ZS-CoSOD~\cite{xiao2024zero} is the training-free method. GCoNet+~\cite{zheng2023gconet+}, VCP~\cite{wang2025visual} and SCoSPARC~\cite{chakraborty2024self} are training-based methods. It's observed that TF-SSD obtains precise co-salient masks compared to other SOTA methods, even in complicated cases (\eg, ``Chopsticks'' and ``Axe'').}
  \label{fig:qualitative}
\end{figure*}

\noindent
\textbf{Qualitative comparison}.
Fig.~\ref{fig:qualitative} shows the visual comparison between TF-SSD and previous SOTA methods. Groups ``Accordion'' and ``Chopsticks'' are from the CoCA dataset. ``Axe'' is from the CoSal2015 dataset and ``Baby\_bed'' is from the CoSOD3k dataset. It's observed that TF-SSD segments co-salient masks that align with the ground truth. Moreover, in complex cases, TF-SSD also predicts accurate masks. For example, in the group of slender characteristics, \ie, ``Chopsticks'', TF-SSD predicts masks with smooth boundaries, while previous methods, no matter whether training-based or training-free methods, fail to accurately segment the co-salient target. In the case of ``Baby\_bed'', TF-SSD also predicts better results than other training-based methods \ie, GCoNet+~\cite{zheng2023gconet+} and VCP~\cite{wang2025visual}. These results demonstrate the strong capability of our TF-SSD.

\subsection{Ablation Study}

\textbf{Effect of proposed components}. Tab.~\ref{tab:ablation_modules} lists comprehensive ablation studies of each component in our TF-SSD. ``QMG-1'' denotes the SAM-based initial filtering, ``QMG-2'' denotes the overlap filtering and ``QMG-3'' denotes the quality assessment in our QMG. We treat the combination of QMG-1 and IPS as our baseline; otherwise, it is impossible to derive the final co-salient predictions. It can be observed that, through the initial filtering, the performance is quite poor, indicating that numerous noisy masks are predicted by SAM. Then, with the help of QMG-2, the performance increases. After adding QMG-3, the first performance leap appears. It yields significant improvements across all evaluation metrics on all datasets. Finally, our ISF pushes the performance to SOTA, showing the importance of the saliency detection ability of DINO. For example, ISF brings a gain of 4.2\% for MAE, 16.4\% for $F_\beta^\text{max}$, 5.0\% for $E_{\xi}^\text{max}$, and 12.9\% for $S_\alpha$ on the CoCA dataset compared to row 3. Overall, they significantly contribute to our TF-SSD.

\begin{table}[!ht]
\centering
	\caption{
		\textbf{Ablation study of the $N-1$ similarity score selection (Eq.~\ref{eq:maxsimilarityscore}) in IPS.} We compare using the maximum similarity (\textit{Max}) versus the average similarity (\textit{Avg}) to compute the similarity score among cross-image prototypes.
	}
    \label{tab:ablation_ipm_strategy_horizontal}
    \small
    \setlength{\tabcolsep}{2pt}
	\begin{tabular}{l|cc|cc|cc}
		\hline 
		\multirow{2}{*}{Strategy} & \multicolumn{2}{c|}{CoCA} & \multicolumn{2}{c|}{CoSal2015} & \multicolumn{2}{c}{CoSOD3k} \\ 
		\hhline{|~|--|--|--|}
		& $F_{\beta}^{\text{max}}\uparrow$ & $E_{\xi}^{\text{max}}\uparrow$ & $F_{\beta}^{\text{max}}\uparrow$ & $E_{\xi}^{\text{max}}\uparrow$ & $F_{\beta}^{\text{max}}\uparrow$ & $E_{\xi}^{\text{max}}\uparrow$ \\
		\hline
        \textit{Avg}
		& 0.568 & 0.778 & 0.775 & 0.826 & 0.737 & 0.792 \\
        \hline
        \textit{\textbf{Max}} 
	    & \textbf{0.686} & \textbf{0.815} & \textbf{0.899}  & \textbf{0.926} & \textbf{0.860}  & \textbf{0.908} \\
		\hline
	\end{tabular}
\end{table}
\noindent
\textbf{Effect of selection strategy in IPS}.
Tab.~\ref{tab:ablation_ipm_strategy_horizontal} compares different selection strategies of the $N-1$ similarity score in Eq.~\ref{eq:maxsimilarityscore}. Our maximum similarity strategy (\textit{Max}) is superior to averaging similarity (\textit{Avg}), which yields a 12.4\% gain in $F_{\beta}^{\text{max}}$ on the CoSal2015 benchmark. This suggests that the maximum similarity exhibits strong semantic correlation across images with the same category.

\begin{table}[!htbp]
\begin{center}
\small
\caption{\textbf{Ablation study on the number of selected masks ($T$) in the ISF module}. Performance is evaluated on the CoCA dataset.}
\begin{tabular}{c|cccc}
\hline
\rule{0pt}{3mm} T & MAE $\downarrow$ & $F_\beta^\text{ max} \uparrow$ & $E_{\xi}^\text{max} \uparrow$ & $S_\alpha \uparrow$\\
\hline
\rule{0pt}{3mm}10 & 0.078 & 0.616 & 0.786 & 0.718 \\
8 & 0.078 & 0.645 & 0.789 & 0.726 \\
7 & 0.077 & 0.666 & 0.809 & 0.744 \\
\textbf{6}  & \textbf{0.077} & \textbf{0.686} & \textbf{0.815} & \textbf{0.763} \\
5 & 0.077 & 0.667 & 0.804 & 0.748 \\
\hline
\end{tabular}
\label{tab:ablation_ail}
\end{center}
\end{table}

\noindent
\textbf{Effect of threshold in ISF}.
Tab.~\ref{tab:ablation_ail} evaluates the number of setected salient masks in our ISF. It's observed that $T=6$ provides the optimal trade-off between candidate recall and noise reduction. Moreover, if we narrow the number of selected masks, it exists over-filtering, while keeping too many masks introduces noise. 
It also suggests that it's worth exploring whether VFMs can provide better salient hints to extract more accurate semantic information for CoSOD.

\section{Conclusion}
In this paper, we introduce a novel method, TF-SSD, to address CoSOD in a training-free manner. First, it employs SAM to acquire a candidate mask pool and introduces a quality mask generator to filter out redundant masks. Then, an intra-image saliency filter is adopted to identify the salient objects within images. Moreover, TF-SSD proposes an inter-image prototype selector that models the co-saliency relations to select the most related masks as the final prediction for CoSOD. Extensive experiments demonstrate that our method achieves significant gains. TF-SSD presents a new paradigm that explores salient semantic understanding upon VFM results, and we envision it serving as a strong baseline to facilitate future research.
\section*{Acknowledgment}
This work was supported by the National Natural Science Foundation of China (No. 62301451, 62301613, 62471405, 62331003), Basic Research Program of Jiangsu (BK20241814), Suzhou Basic Research Program (SYG202316), XJTLU REF-22-01-010 and XJTLU RDF-22-02-066.


\section{Supplementary Materials}
This supplementary material provides additional details to complement our main paper. We elaborate on our implementation details in Sec.~\ref{sec:1}, present more quantitative and qualitative results in Sec.~\ref{sec:2} and Sec.~\ref{sec:3}, and discuss the limitations of our method in Sec.~\ref{sec:4}.

\section{Additional implementation details}
\label{sec:1}
To handle challenging cases, several mechanisms are necessary to ensure robustness in practice. Due to space limitations in the main paper, we describe these details below.

\subsection{Fallback mechanism in ISF}
Our Intra-image Saliency Filter (ISF) relies on DINO's attention maps to identify salient objects from SAM~\cite{sam} proposals. However, in some challenging scenarios, QMG may fail to segment salient masks, where all masks yield low saliency scores $s_{n,t}^{sal}$. To ensure robustness, we employ a fallback mechanism that directly generates masks from attention maps to avoid these bad cases.

In particular, if the maximum saliency score $\max_t(s_{n,t}^{sal})$ is lower than a threshold $\tau_{fb}$, the attention map $\mathcal{A}_n$ is binarized to obtain a mask $m_{n,t}^{fb}$ that captures the salient region, formulated as:

\begin{equation}
m_{n,t}^{fb}(x,y) =
\begin{cases}
1 & \text{if } A_n(x,y) > \tau_{attn},\\
0 & \text{otherwise},
\end{cases}
\label{eq:fallback_mask:supp}
\end{equation}

where $\tau_{attn}$ is the binarization threshold for $\mathcal{A}_n$. This single fallback mask $m_{n,t}^{fb}$ will replace the low-quality masks. This operation ensures that each image retains at least one salient candidate mask for subsequent processing. In our implementation, we set $\tau_{fb}=0.05$, and $\tau_{attn}$ is dynamically set to a threshold that retains the top 50\% of the highest attention values for each map $\mathcal{A}_n$.

\subsection{Merging of multiple co-salient objects in IPS}

Our Inter-image Prototype Selector (IPS) is designed to identify the target co-salient object in each image. However, individual images may contain multiple instances of the co-salient object, and SAM often segments them separately. To obtain accurate results for CoSOD, we have to merge these separate masks into a single one.

To address such cases, we employ a dual-verification mechanism with two thresholds: a semantic similarity threshold $\tau_{sem}$ and a consistency difference threshold $\tau_{diff}$. For each image $I_n$ with a selected mask $\hat{m}_n$, we examine the remaining masks in $\mathcal{M}_n^{salient}$ to identify extra co-salient objects. A candidate mask $m_j$ is considered an extra co-salient object if it satisfies the following two conditions:

\begin{enumerate}[1)]
    \item \textbf{Semantic similarity:} The candidate mask is semantically similar to the primary mask, measured by pairwise similarity $\langle p_{\hat{m}_n}, p_j \rangle \geq \tau_{sem}$.
    \item \textbf{Cross-image consistency:} The candidate mask exhibits comparable cross-image consistency, measured by $|S_j^{co} - S_{\hat{m}_n}^{co}| < \tau_{diff}$.
\end{enumerate}

The first condition ensures semantic coherence within the image, while the second verifies that the candidate mask represents an object that appears consistently across the group. Verified masks are merged with the primary mask to form the final segmentation mask. In our implementation, $\tau_{sem}$ is dynamically set as the top 80\% of intra-image pairwise similarities, and $\tau_{diff}$ is set to 0.1.

\section{Additional quantitative results}
\label{sec:2}

\subsection{Different SAM area ratios}

The area ratio threshold $\tau_{area}$ is a critical parameter in our QMG that filters out excessively small masks (Eq.~2 in the main paper). To evaluate its impact on CoSOD performance, we conduct ablation experiments with four different threshold values: 0.002, 0.005, 0.01, and 0.02 on the CoCA~\cite{fan2020taking} dataset. We employ two metrics: (1) \textbf{F-measure}: the final CoSOD performance obtained using the complete TF-SSD framework; (2) \textbf{Oracle F}: the upper bound performance computed using selected oracle masks from QMG.

As shown in Tab.~\ref{tab:area_ratios}, when the threshold is too low (0.005), excessive trivial masks are retained in the candidate set, introducing noise that degrades the subsequent ISF and IPS. 
Meanwhile, the Oracle F score reflects the quality of the candidate pool. When the threshold is too high (0.02), some valid co-salient objects are incorrectly filtered out, which limits both the quality of the candidate pool and the final performance. $\tau_{area}=0.01$ strikes an optimal balance, which effectively removes trivial masks while preserving diverse co-salient object candidates.

\begin{table}[t!]
\centering
\caption{Ablation study on different area ratio thresholds $\tau_{area}$ on the CoCA dataset.}
\label{tab:area_ratios}
\small
\begin{tabular}{c||c|c}
\hline
\rule{0pt}{3mm}
$\tau_{area}$ & $F_{\beta}^{\text{max}}$ & Oracle $F_{\beta}^{\text{max}}$ \\
\hline
\rule{0pt}{3mm}0.002 & 0.545 & 0.696 \\
0.005 & 0.601 & 0.722 \\
\rowcolor{mygray}\rule{0pt}{3mm}\textbf{0.01} & \textbf{0.686} & \textbf{0.768} \\
0.02 & 0.653 & 0.737 \\
\hline
\end{tabular}
\end{table}

\subsection{Performance using other DINO backbones}

Our main experiments use DINO ViT-B/8~\cite{dino} as the backbone for feature extraction in ISF. To provide a broader performance assessment, we compare it with two other notable backbones: (1) ViT-B/16~\cite{dino} from the same DINO framework with a larger patch size; (2) ViT-B/14~\cite{oquab2023dinov2} from the more recent DINOv2 framework. We conduct experiments on the CoCA dataset, which highlights the trade-offs between different DINO versions.

As shown in Tab.~\ref{tab:dino_backbones}, DINO ViT-B/8 significantly outperforms other configurations across all three metrics. A smaller patch size enables our model to capture finer spatial details, which is crucial for feature matching and prototype selection in our ISF and IPS. In contrast, although DINOv2 ViT-B/14 utilizes more advanced pretraining strategies, its larger patch size limits the perception of fine-grained information. This result demonstrates that a smaller patch size is more suitable for CoSOD that require precise localization and fine-grained feature alignment.

\begin{table}[ht]
    \centering
    \small
    \caption{Performance comparison of different DINO backbones on the CoCA dataset.} 
    \setlength{\tabcolsep}{4mm}\begin{tabular}{l||c|c|c}
        \toprule
        Backbone         & $F_{\beta}^{\text{max}}$      & $E_{\xi}^{\text{max}}$      & $S_{\alpha}$\\
        \midrule
        DINO ViT-B/16   &0.556              & 0.742             &0.697 \\
        DINOv2 ViT-B/14           &0.607               &0.783           &0.708  \\
        \rowcolor{mygray}
        \textbf{DINO ViT-B/8}        & \textbf{0.686}              & \textbf{0.815}  & \textbf{0.763} \\
        \bottomrule
    \end{tabular}
    \label{tab:dino_backbones}
\end{table}

\subsection{Ablation on hyperparameter settings}

We conduct additional ablation studies to evaluate the impact of hyperparameter settings.

\noindent\textbf{Quality score weights.}
The balanced quality score $S_{n,t}^{ba}$ (Eq.~5 in the main paper) combines IoU prediction confidence and area preference through weights $\alpha$ and $\beta$, where $\alpha + \beta = 1$ to ensure normalized weighting between the two terms. Tab.~\ref{tab:alpha_beta} illustrates the performance with different weight settings on the CoCA dataset.

\begin{table}[ht]
    \centering
    \small
    \caption{Ablation on quality score weights on the CoCA dataset.}
    \setlength{\tabcolsep}{2.5mm}\begin{tabular}{cc||c|c|c}
        \toprule
        $\alpha$ & $\beta$ & $F_{\beta}^{\text{max}}$ & $E_{\xi}^{\text{max}}$ & $S_{\alpha}$\\
        \midrule
        0.0 & 1.0 & 0.473 & 0.702 & 0.611 \\
        0.3 & 0.7 & 0.591 & 0.737 & 0.651 \\
        0.5 & 0.5 & 0.658 & 0.795 & 0.744 \\
        \rowcolor{mygray}
        \textbf{0.7} & \textbf{0.3} & \textbf{0.686} & \textbf{0.815} & \textbf{0.763} \\
        1.0 & 0.0 & 0.641 & 0.782 & 0.735 \\
        \bottomrule
    \end{tabular} 
    \label{tab:alpha_beta}
\end{table}

\noindent \textbf{Ideal size range.}
The ideal size range $[r_{min}, r_{max}]$ (Eq.~4 in the main paper) defines the appropriate size for co-salient objects. Tab.~\ref{tab:size_range} presents results for different range settings. The range $[0.15, 0.7]$ performs best, suggesting that co-salient objects typically occupy 15\%-70\% of the image area. More restrictive ranges (\eg [0.2, 0.5]) limit the number of valid masks, while much wider ranges (\eg [0.05, 0.85]) fail to effectively filter out background and large masks.

\begin{table}[ht]
    \centering
    \small
    \caption{Ablation on ideal size range on the CoCA dataset.}
    \setlength{\tabcolsep}{2.5mm}\begin{tabular}{c|c||c|c|c}
        \toprule
        $r_{min}$ & $r_{max}$ & $F_{\beta}^{\text{max}}$ & $E_{\xi}^{\text{max}}$ & $S_{\alpha}$\\
        \midrule
        0.05 & 0.85 & 0.661 & 0.798 & 0.747 \\
        0.1 & 0.6 & 0.678 & 0.808 & 0.756 \\
        \rowcolor{mygray}
        \textbf{0.15} & \textbf{0.7} & \textbf{0.686} & \textbf{0.815} & \textbf{0.763} \\
        0.2 & 0.5 & 0.619 & 0.741 & 0.669 \\
        \bottomrule
    \end{tabular} 
    \label{tab:size_range}
\end{table}

\begin{table}[ht]
    \centering
    \small
    \caption{Ablation on overlap threshold on the CoCA dataset.}
    \setlength{\tabcolsep}{3mm}\begin{tabular}{c||c|c|c}
        \toprule
        $\tau_{con}$ & $F_{\beta}^{\text{max}}$ & $E_{\xi}^{\text{max}}$ & $S_{\alpha}$\\
        \midrule
        0.5 & 0.652 & 0.791 & 0.741 \\
        0.7 & 0.673 & 0.805 & 0.754 \\
        \rowcolor{mygray}
        \textbf{0.85} & \textbf{0.686} & \textbf{0.815} & \textbf{0.763} \\
        0.95 & 0.679 & 0.796 & 0.756 \\
        \bottomrule
    \end{tabular} 
    \label{tab:overlap}
\end{table}

\noindent\textbf{Overlap threshold.}
The overlap threshold $\tau_{con}$ controls how aggressively overlapping masks are filtered. As shown in Tab.~\ref{tab:overlap}, $\tau_{con}=0.85$ achieves the optimal performance. Lower thresholds (\eg, 0.5) retain too many redundant masks, while higher thresholds (\eg, 0.95) will incorrectly remove valid masks that partially overlap with larger objects. They both lead to the reduced diversity of candidate masks for subsequent processing.

\section{Additional qualitative results}
\label{sec:3}
In this section, we present visualizations of several key stages in our TF-SSD framework. 

\begin{figure*}[!t]
  \centering
  \includegraphics[width=0.95\textwidth]{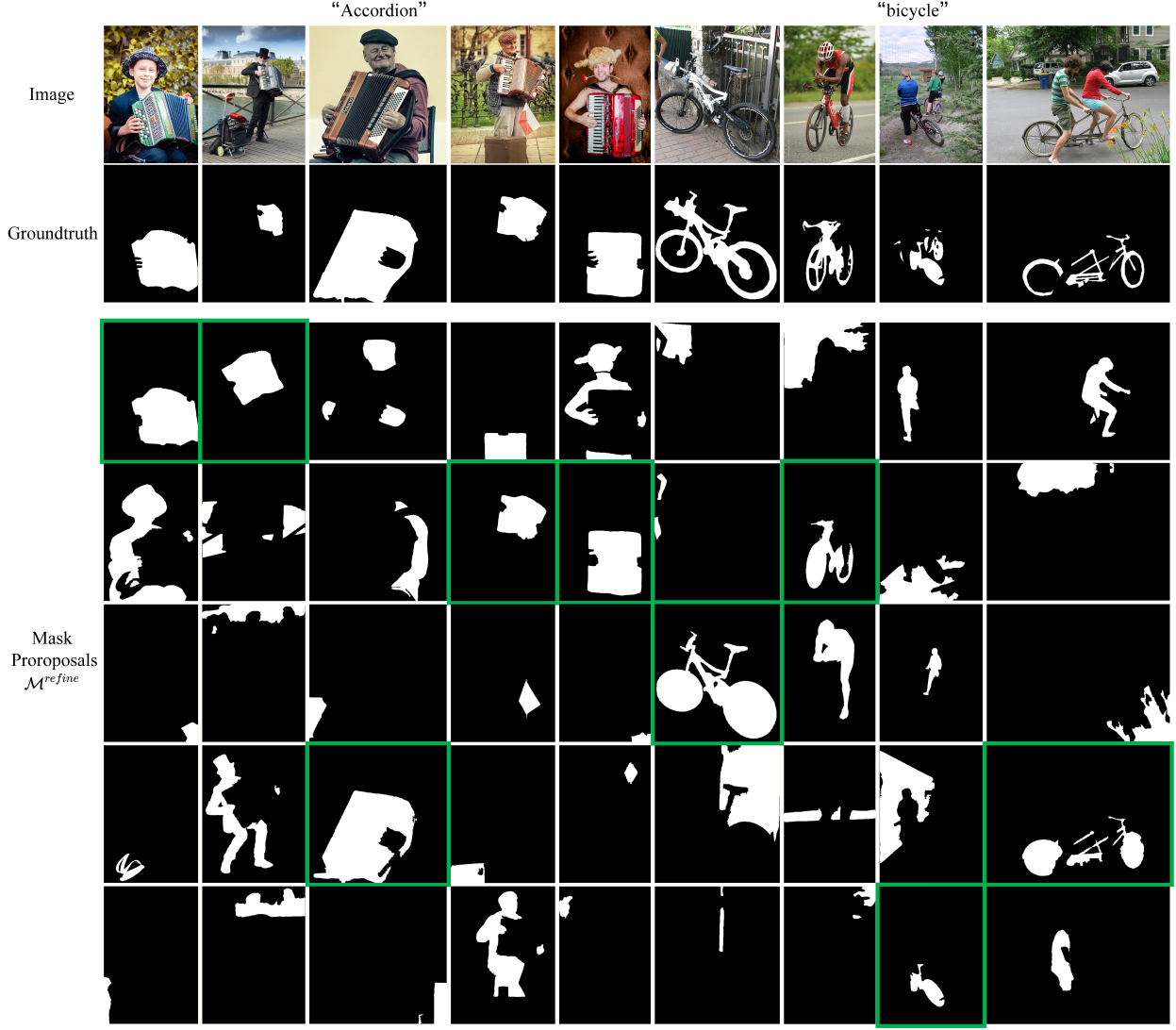}  
  \caption{Visualization of SAM mask proposals for "Accordion" and "bicycle" image groups. For each group, we show the original images (row 1), ground truth(GT) masks (row 2), and the top-5 mask proposals ranked by quality score $S_{n,t}^{ba}$ (rows 3-7). Green borders indicate the final predictions selected by our TF-SSD pipeline.}
  \label{fig:1}
\end{figure*}

\subsection{Visualization of SAM proposals}
To demonstrate SAM's segmentation capability and the effectiveness of our QMG module, Fig.~\ref{fig:1} visualizes the top-5 mask proposals ranked by our quality score $S_{n,t}^{ba}$ (Eq.~5 in the main paper). The masks with green borders represent our final predictions after the complete TF-SSD pipeline. It's observed that SAM can generate mask proposals that closely align with the GT, and our quality score $S_{n,t}^{ba}$ effectively ranks these masks at the top positions. 
It validates the effectiveness of SAM for mask generation and confirms that QMG can successfully identify promising candidate masks for subsequent processing.

\subsection{Visualization of DINO attention maps}

To illustrate the saliency-aware capability of DINO's attention mechanism, Fig.~\ref{fig:2} visualizes the attention maps $\mathcal{A}_n$ of DINO's CLS-token. The attention maps can highlight the salient object regions that closely align with the GT masks. These observations validate the effectiveness of leveraging DINO attention maps to select salient mask proposals from the refined candidate set derived from our QMG.

\section{Limitations}
\label{sec:4}

While our TF-SSD framework achieves strong performance on most CoSOD benchmarks, it still exhibits limitations in detecting small co-salient objects. 
As illustrated in Fig.~\ref{fig:limitations}, the "moon" category contains predominantly small objects that occupy a relatively small portion of the image. Due to SAM's initial area-based filtering mechanism in our QMG, masks with small area ratios ($r_{n,t}^{area} < \tau_{area}$) are typically denoted as trivial objects that are filtered out. In addition, small objects also tend to receive lower quality scores $S_{n,t}^{ba}$ from DINO's attention maps, which leads to wrong identification for co-salient objects. 

Our future work concentrates on addressing these bad cases in challenging scenarios. We will explore incorporating semantic cues into an adaptive threshold that is modulated according to contexts of the image. We argue that it can identify the small salient objects from the trivial ones.

\begin{figure*}[!htbp]
  \centering
  \includegraphics[width=0.99\textwidth]{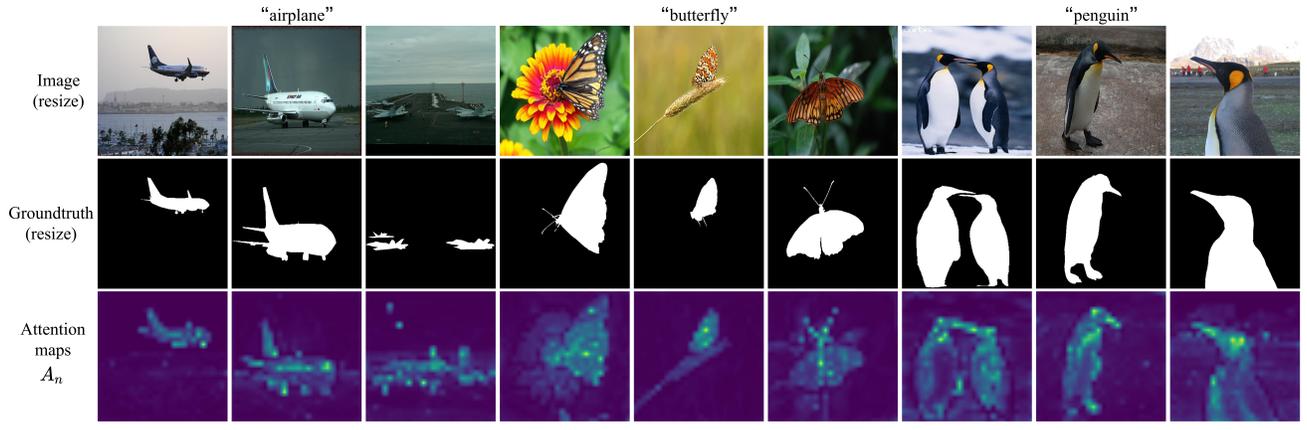}  
  \caption{Visualization of DINO attention maps for three categories: "airplane", "butterfly", and "penguin". Row 1: Resized input images. Row 2: Resized GT masks. Row 3: DINO attention maps $\mathcal{A}_n$ that naturally highlight salient objects.}
  \label{fig:2}
\end{figure*}

\begin{figure}[!htbp]
  \centering
  \includegraphics[width=\columnwidth]{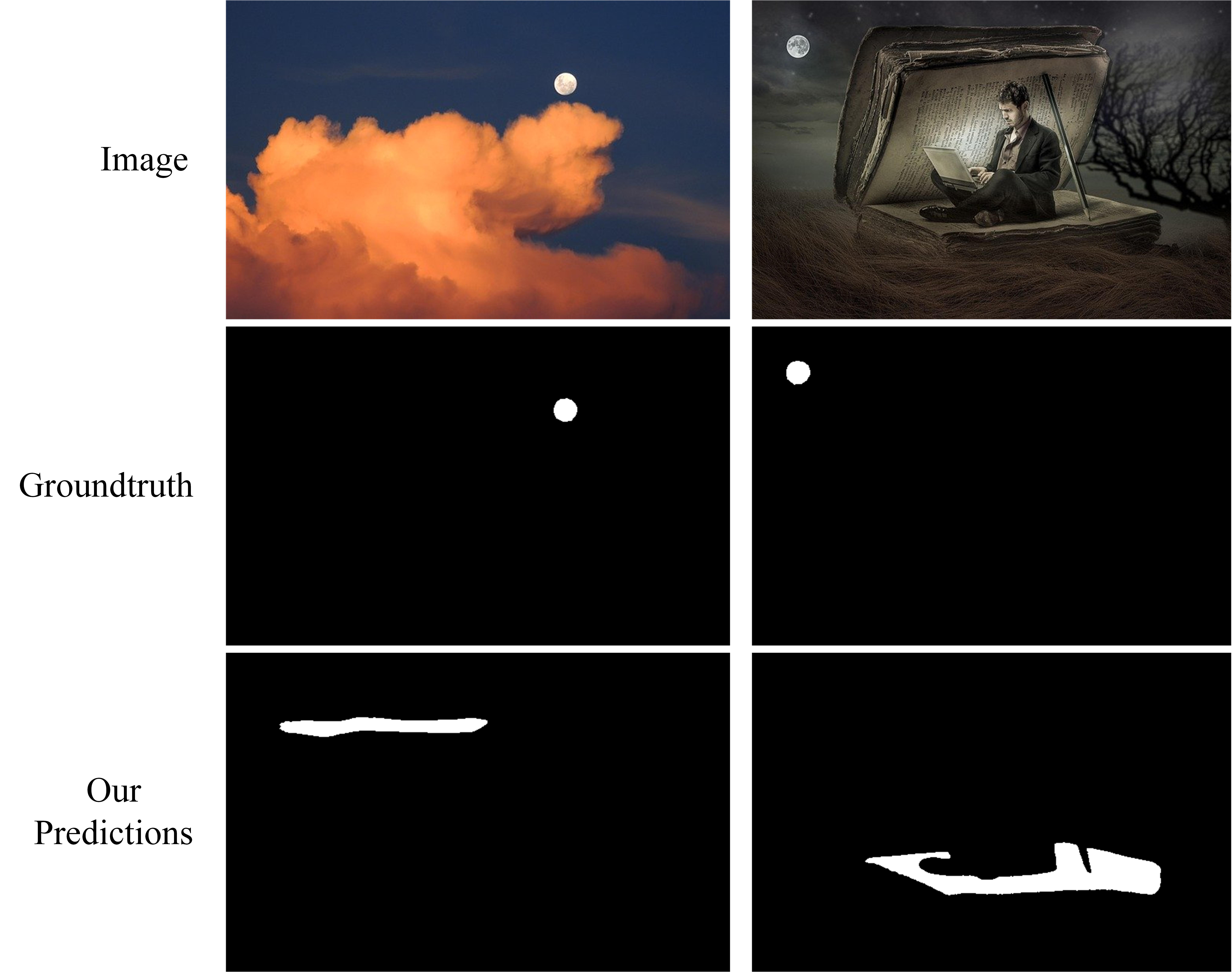}  
  \caption{Limitation on small object detection. Visualizations of failed cases from the "moon" category.}
  \label{fig:limitations}
\end{figure}



\end{document}